\documentclass[11pt,a4paper]{article}

\usepackage[utf8]{inputenc}
\usepackage[T1]{fontenc}
\usepackage[USenglish]{babel}
\usepackage{lmodern}
\usepackage[margin=2.5cm]{geometry}
\usepackage{amsmath,amssymb}
\usepackage{graphicx}
\usepackage{microtype}
\usepackage{authblk}
\usepackage[hidelinks]{hyperref}

\title{Evaluating the Transfer of Co-Evolved Communication from 2D to 3D Simulation}

\author{Fernando Montes-Gonzalez}
\affil{Instituto de Investigaciones en Inteligencia Artificial,\\
Universidad Veracruzana, Mexico\\
\texttt{fmontes@uv.mx}}

\date{}

\begin{document}

\maketitle

\begin{abstract}
This work examines the transfer of a co-evolved communication mechanism
between two robotic agents from a discrete two-dimensional (2D)
simulator to a three-dimensional simulator with real physics (3D). The
study focuses on whether a communication mechanism co-evolved in a 2D
environment retains its functional role after transfer to a 3D
physics-based simulator. To support this analysis, the effects of the
episode time budget, the social cue, and the asymmetry between the two
co-evolved roles were examined.

The results indicate that the success rate increased approximately
linearly with the evaluated time budgets, with no evidence of a plateau
between 2,000 and 6,000 physics steps, suggesting that evaluations
based on shorter episodes may underestimate the performance of the
trained controllers. In both simulators, the social cue functioned
primarily as a jam-assistance mechanism rather than as a navigation
guide, although with a more pronounced effect in 2D. Analysis of eight
independent evolutionary runs revealed a consistent direction of
asymmetry, although its magnitude varied across runs. Controlling the
processing order between agents allowed us to rule out an artifact of
the physics engine. Finally, the results are discussed in terms of the
factors that may contribute to the remaining performance gap observed
after transfer.
\end{abstract}


\section{Introduction}
\label{sec:introduction}

Communication between robotic agents can emerge without explicit design
when artificial evolution selects for control mechanisms capable of
coordinating behavior through signals \cite{nolfiEvolutionaryRoboticsBiology2000a}. Previous work in
evolutionary robotics has shown that once a basic signal-response system
is established in a population, it stabilizes and becomes more complex
through selection, even when the fitness function does not explicitly
reward communication itself \cite{maroccoOriginsCommunicationEvolving2006}. Furthermore, the functional content
of an evolved signal may not align with the intuitive interpretation of
its purpose. A signal introduced to encourage agents to approach each other may ultimately function as a stall-assistance mechanism rather than as a general navigation guide.

One question that has received less attention is whether a communication mechanism evolved under simplified simulation conditions retains its function after transfer to a simulator with real-world physics. This problem falls within the
broader discussion of the ``reality gap'' in evolutionary robotics.
Controllers optimized in a simulator can exploit simulator regularities
that do not hold true in a more realistic physical environment, even
when the simulator reasonably accurately captures the task to be solved
\cite{jakobiNoiseRealityGap1995}. In their work, the authors \cite{trianniSelforganisationCommunicationGroups2006} compared self-organized
communication between simulated and physical robots and found that the
mechanism does not always transfer with the same efficiency. Recently, the improvement of signal communication in a foraging task was studied using evolutionary robotics \cite{aldana-francoImprovementSignalCommunication2024}. A related study analyzed the value of information and the emergence of communication in co-evolved e-puck robots within a two-dimensional environment \cite{MontesGonzalez2026ValueInformation}.

This paper examines the transfer of a co-evolved communication mechanism previously studied in a two-dimensional, discrete simulation engine without real-world physics \cite{MontesGonzalez2026ValueInformation} and subsequently evaluated in a three-dimensional simulator with friction, collision, and inertia. The analysis focuses on three aspects of this transfer: the influence of episode duration on performance, the functional role of the social signal, and the asymmetry observed between the two co-evolved agents.

The remainder of this work is organized as follows. Section~\ref{sec:methods} describes the two simulation engines, the controller architecture, the social signal and fitness function, the stall criterion, the basal ganglia-inspired selective layer, the evolutionary procedure, and the proposed diagnostic tools. Section~\ref{sec:results} presents the performance baseline, the relationship between budget and performance, the effect of the signal on stall resolution, and the weight asymmetry between agents. Section~\ref{sec:discussion} integrates the findings and discusses possible factors associated with the observed transfer performance and agent asymmetry. Finally, the limitations of the work are noted, and avenues for future research are proposed.

\section{Methods}
\label{sec:methods}

\subsection{Simulation Engines}

\begin{figure}[!ht] \centering \begin{minipage}{0.48\textwidth} \centering \includegraphics[width=\linewidth]{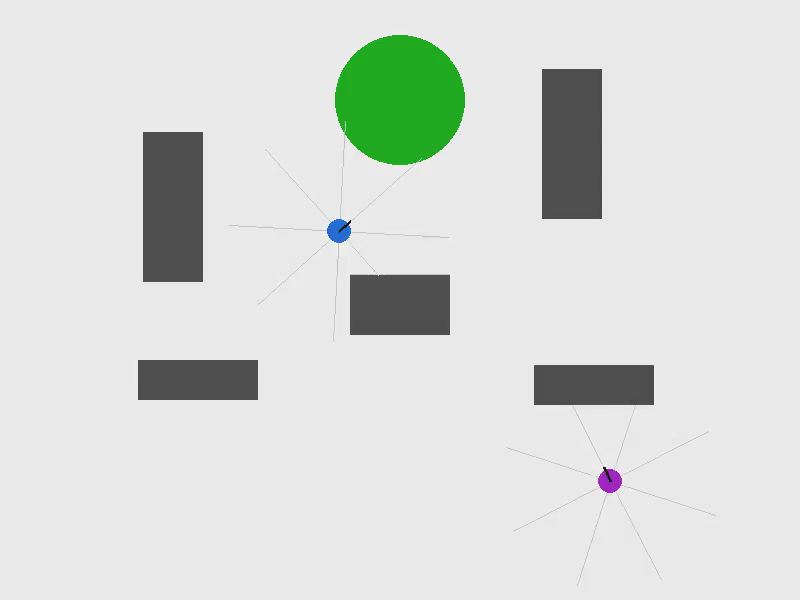} \vspace{0.3em} (a) \end{minipage} \hfill \begin{minipage}{0.48\textwidth} \centering \includegraphics[width=\linewidth]{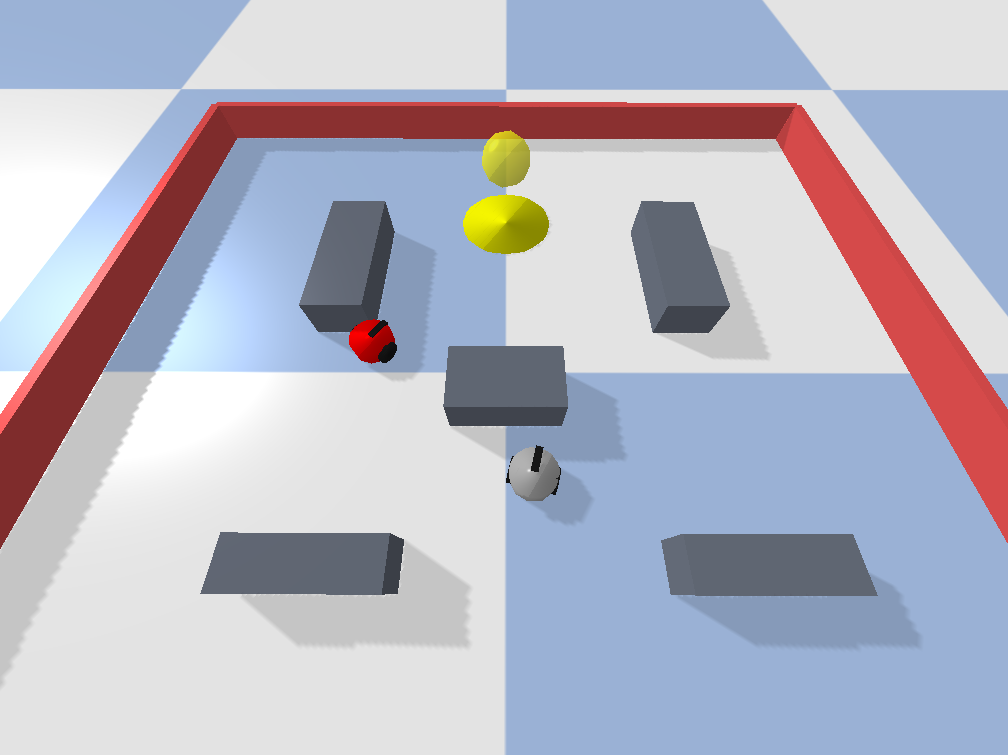} \vspace{0.3em} (b) \end{minipage} \caption{ Comparison of the simulation engines used for the evaluation of coevolved communication between robotic agents. (a) 2D
engine with discrete environment representation, without physical
modeling, used for low computational-cost simulations. (b) 3D engine
based on PyBullet, with 240 Hz physical simulation and decoupled
simulation time control through decision updates every 15 physics steps
($\approx$16 Hz), equivalent to a control cycle commonly used in robotic
simulators. Both environments contain two agents, shared obstacles, and
shared objectives for performing cooperative tasks. } \label{fig:sim_engines} \end{figure}

Two simulation engines were used to evaluate co-evolved communication
between two robotic agents (called \emph{brain\_a} and \emph{brain\_b}). The
two-dimensional engine implements an environment based on a discrete
grid without real physics (Figure~\ref{fig:sim_engines}a). The three-dimensional engine implements a physical model using PyBullet \cite{CoumansBai2016}, with friction, collision, and inertia between the simulated bodies (Figure~\ref{fig:sim_engines}b). In the 3D
environment, each agent has its own independent random number generator.
This difference is relevant when analyzing potential effects associated
with the processing order of the agents.

In the 3D engine, physics is simulated at 240 Hz (4.17 ms timestep in
PyBullet). The controller makes decisions every 15 physics steps
(\emph{brain\_ratio=15}), equivalent to 62.5 ms of simulated time per
decision ($\approx$16 Hz). The decision interval is of an order of magnitude comparable to that of a typical control cycle in robotics simulators such as Webots \cite{michelWebotsSymbiosisVirtual1998}, whose default time step is 32 ms, with 64 ms as a common alternative value. The total budget of an episode is expressed in physics steps (\emph{steps}); the effect of this
parameter on performance is characterized in Section~\ref{sec:results}.

\begin{figure}[!ht] \centering \includegraphics[width=0.4\textwidth]{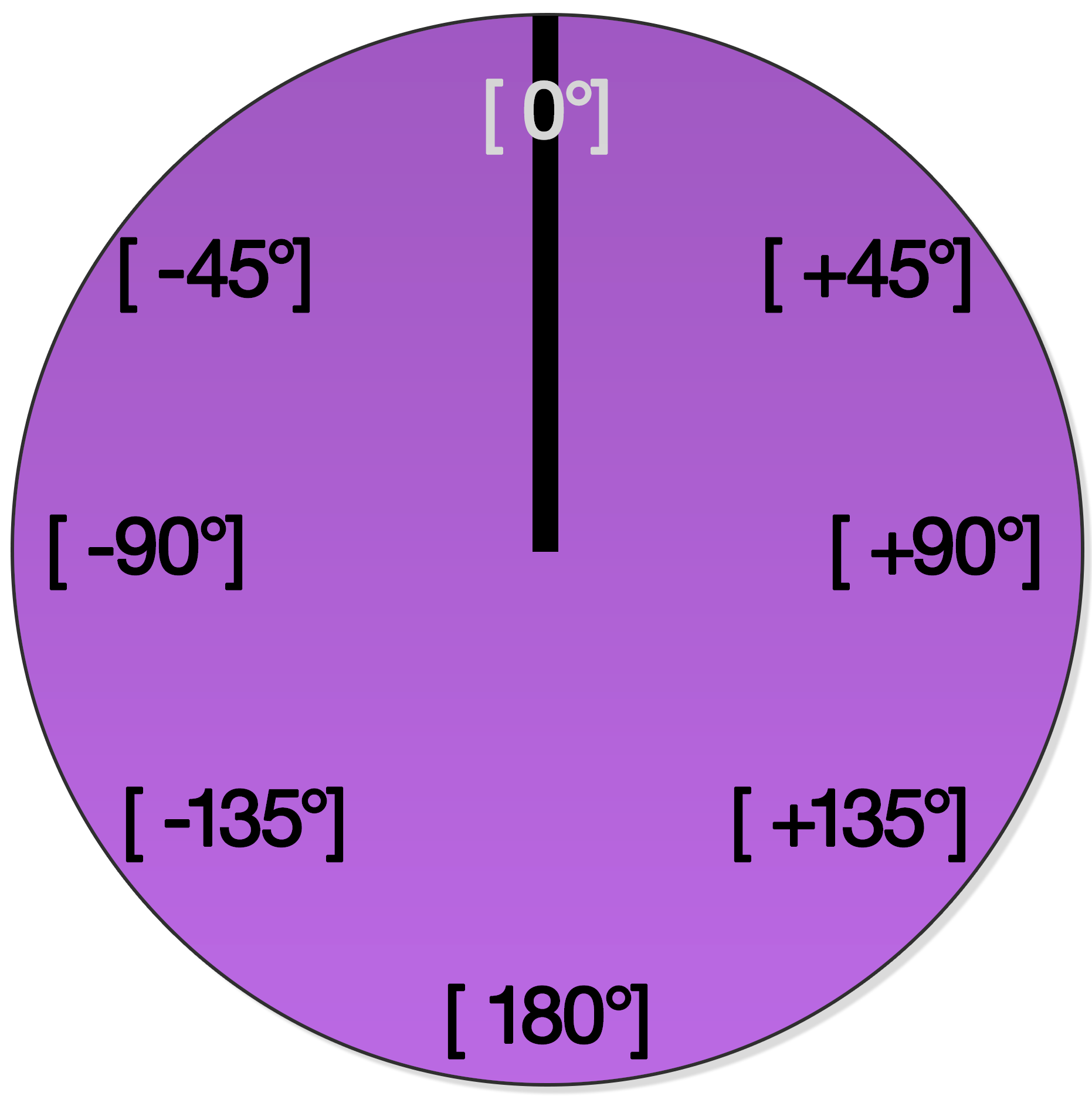} \caption{ Distribution of the robot's eight proximity sensors. The sensors are evenly spaced at 45$^\circ$ intervals around the robot, providing 360$^\circ$ detection coverage. The black line indicates the robot's forward-facing direction and serves as a visual reference for interpreting sensor orientation in both the 2D and 3D simulators. } \label{fig:proximity_sensors} \end{figure}

The robot morphology used in this study is inspired by the e-puck robot \cite{Mondada2009}. Figure~\ref{fig:proximity_sensors} summarizes only the features relevant to the interpretation of the methods and results presented. Each
robot's body is equipped with eight proximity sensors
arranged symmetrically in a ring around its perimeter, spaced 45° apart,
providing 360° detection coverage. Each sensor obtains its reading via a
ray drawn from the robot's center, with a maximum range
of 0.20 meters. The returned value ranges from 0.0, when there are no
obstacles within that range, to 1.0, when the obstacle is in contact
with the robot's surface. Rays that intersect the
robot's own body are excluded from the reading. For locomotion, each robot has wheels with a radius of 0.02 meters and a maximum linear speed of 0.15 m/s, selected to maintain stable navigation within the dimensions of the simulated environment.

\subsection{Controller Architecture}

Each agent is controlled by a recurrent network of the Gated Recurrent
Unit (GRU) type \cite{choLearningPhraseRepresentations2014}, co-evolved independently for each role
(\emph{brain\_a} and \emph{brain\_b}). The network combines a recurrent path, with internal memory, and a direct residual path between the input and output logits without going through the GRU recurrent memory.

The output layer (\emph{w\_out}) produces the logits for the five discrete actions of the controller (\emph{move\_forward}, \emph{left\_turn}, \emph{right\_turn}, \emph{move\_backward}, \emph{signal/explore}), from which the action to be executed is selected using argmax. This work focuses specifically on the columns corresponding to the turns (left\_turn, right\_turn), as these support the structural bias analysis between agents reported in Section~\ref{sec:3.4}. In the 3D simulator, each discrete action is subsequently translated into a pair of angular wheel velocities for the differential-drive robot (\texttt{move\_forward}: [1.0, 1.0]; \texttt{left\_turn}: [-0.3, 0.6]; \texttt{right\_turn}: [0.6, -0.3]; \texttt{move\_backward}: [-0.5, -0.5]; \texttt{signal/explore}: [0.8, 0.8]), which are executed through differential-drive control during the interval between decisions. These values were selected empirically to provide stable locomotion and turning behavior within the dimensions and obstacle density of the simulated environment. The same action mapping was used throughout all experiments. The left\_turn and right\_turn velocity pairs are unequal-magnitude pivots between the two wheels, designed to generate rotation with minimal net translation. This property is relevant for interpreting the turning behavior observed in the agents.

The residual pathway (\emph{w\_res}) contributes directly to those same
logits from the input vector, without going through the
GRU's recurrent memory. An amplification parameter
(\emph{SOCIAL\_RESIDUAL\_BOOST}) specifically scales the contribution of
the social channel ($x[10]$) within this residual pathway, leaving the
rest of the weights and memory dynamics unchanged.

With SOCIAL\_RESIDUAL\_BOOST = 1.0, this contribution remains unchanged:
the social channel carries the same weight as if the parameter did not
exist. A direct diagnostic of the already trained weights showed that,
with this value, the social channel had a very weak influence on the
controller's decision compared to the other inputs. For
this reason, the official checkpoint reported in this work
(\emph{v15\_full\_final}) was trained with SOCIAL\_RESIDUAL\_BOOST =
3.0, a value chosen to amplify that influence without altering the
controller architecture.

Within the controller's input vector, two components are
relevant to this work: $x[8]$, which encodes the direction to the food, and $x[10]$, the social channel, described in the following section (Section~\ref{sec:social_signal}).

\subsection{Social signal and fitness incentive}
\label{sec:social_signal}

The social channel ($x[10]$) is defined as the product of the relative angle to the partner ($\theta$) and an exponentially decaying distance factor based on the distance to the partner ($d$): \begin{equation} x[10]= (\theta/\pi)\, e^{-d/\mathrm{SOCIAL\_SIGNAL\_DECAY}} \end{equation} The signal is active only when the partner has already reached the food and is signaling; otherwise, $x[10] = 0$. The decay constant (\textit{SOCIAL\_SIGNAL\_DECAY} = 0.563) was empirically calibrated as the median observed recruitment distance, such that the signal retains approximately 37\% of its strength at that representative distance. An additional parameter (\emph{recruit\_prob}) determines the probability that an agent that has already reached the food will actually signal at each decision step instead of remaining silent. In the official checkpoint used for the results reported in this work, \emph{recruit\_prob} was set to 0.90.

The fitness incentive combines four components. Formally, the fitness accumulated per episode for agent $i$ can be expressed as: 
\begin{equation} \begin{split} F_i=\sum_{t=1}^{T} \Big( &\Delta d_{food}(t)K_{dist} +1_{stalled}K_{stalled} \\ &+1_{rescue}B_{rescue} +1_{recruit}B_{recruit} \\ &-1_{abandon}P_{abandon} \\ &+1_{success}B_{terminal}\mu(gap) \Big) \end{split} \label{eq:fitness} \end{equation}
In Equation~\ref{eq:fitness}, $\Delta d_{food}(t)$ denotes the net approach to the food at decision step $t$. The first term, weighted by $K_{dist}$ (\textit{DIST\_MULT} = 150,000), is applied at each of the approximately 133 decisions per episode. The second term applies a penalty ($K_{stalled} = -100$) whenever the agent is classified as stalled. The third term grants a social bonus ($B_{rescue} = 20,000$) when the agent approaches its partner while the partner is signaling from the food location, provided that the approach toward the partner strictly exceeds the agent's own approach toward the food within the same time window. This condition prevents the bonus from being obtained through simple geometric coincidence. The fourth term applies once the agent has already reached the goal. While the partner has not yet arrived, a bonus ($B_{recruit} = 750$) is awarded whenever the agent signals, whereas a penalty ($P_{abandon} = 2,500$) is applied whenever it does not. Finally, upon successful completion of the episode, a terminal reward ($B_{terminal} = 3,000,000$) is added, scaled by a factor $\mu(gap)$, which equals 1.0 when the arrival gap between the two agents is smaller than \textit{TERMINAL\_GAP\_THRESHOLD}, and 0.3 otherwise. 

The large difference in magnitude between the terminal reward and the auxiliary rewards is intentional and follows the reward-shaping paradigm \cite{ngPolicyInvarianceReward1999}. The auxiliary components guide the search process toward conditions associated with successful task completion rather than competing with the terminal reward itself. As discussed in Section~\ref{sec:3.3}, recalibrating \textit{SOCIAL\_RESCUE\_BONUS} from 500 to 20,000 produced observable behavioral changes despite remaining more than two orders of magnitude smaller than the terminal reward.

Neither of the social components, such as the approach bonus or the recruitment bonus/penalty, are explicitly conditioned on the receiving agent being stuck. Instead, they function as general rewards, defined in terms of distance and signal persistence for the approach bonus and recruitment bonus, respectively.
Their specific manifestation as a stall-resolution mechanism is an emergent result of training, not a property imposed by the fitness function design.

The value of $B_{rescue}$ (\textit{SOCIAL\_RESCUE\_BONUS}) used in the official checkpoint is 20,000, resulting from a recalibration of the original value (500), which was several thousand times smaller than the terminal reward and was smaller than the distance-shaping reward associated with a single navigation decision. The values of $B_{recruit}$ and $P_{abandon}$ (750 and -2,500, respectively) preserve the same relative proportion as in the original formulation of the system and were not specifically recalibrated in this work.

\subsection{Stall criterion}
\label{sec:stall_criterion}

In the 3D engine, an agent is considered stuck (\emph{stalled}) when it
commands movement but its effective displacement remains below a minimum
threshold of \emph{displacement $< 0.0001$} during 25 consecutive
physics steps. This activation threshold (25 steps and a displacement threshold of 0.0001) was inherited from the 2D environment without recalibration, as its intended function is simply to distinguish effective movement from blockage.

The stalled state is deactivated, instead, by a tolerant window criterion
specifically calibrated in this work: of the last 30 physics steps, the
agent is considered freed as soon as 24 of them correspond to actual
movement above the threshold, without requiring them to be consecutive.
The value of 24 out of 30 was obtained by instrumenting the internal release counter under the original criterion, which required a perfect streak of 30 consecutive movement ticks. During these observations, agents repeatedly reached 28 or 29 successful ticks out of 30 before a single unfavorable tick reset the counter to zero, generating artificially long stalled episodes even though
the agent was already effectively moving most of the time. This
criterion, defined in the simulation environment (\emph{foraging\_env3d.py}),
is the one used in all stall-resolution diagnostics reported in Section~\ref{sec:results}.

\subsection{Basal ganglia as an action selection mechanism}

In addition to the policy learned by the GRU controller, each agent incorporates a fixed layer of instinct rules that acts as a basic unevolved action selector. This mechanism was inherited from the 2D implementation and can override or condition the action selected by the network in three specific situations.

\textbf{Recruitment}: If the agent has already reached the food, with
probability recruit\_prob the action is forced to signal instead of
using the network output, leaving the rest of the decisions free for the
network to decide for itself (see Section~\ref{sec:social_signal}).

\textbf{Stall assistance}: If the agent is classified as stuck (criterion defined in Section~\ref{sec:stall_criterion}), the selected action is replaced by \emph{signal/explore} with a base probability of 0.20. This probability remains at 0.20 for up to eight consecutive stalled decisions and increases linearly from the ninth onward to a maximum of 0.95, never reaching 1.0. This preserves some behavioral variability even during prolonged stalls. The streak counter is reset as soon as the agent is no longer classified as stuck.

\textbf{Avoid fixation} (\emph{Annealing}): If the network selects \emph{signal/explore} through its own policy rather than through an instinct rule, the action is replaced with a random navigation action with probability \emph{h\_prob}. This probability follows a decreasing schedule throughout evolution, reaching zero around generation 80, and is active only during training. For the checkpoints reported in this work, \emph{h\_prob} is fixed at zero.

The three mechanisms were retained as fixed rules, consistent with the original 2D implementation. They provide basic exploration and recovery behaviors from which evolution can develop more specialized navigation and communication strategies. This layer is conceptually motivated by the work of \cite{prescottSimulatedDopamineModulation2024b}, who studied action selection in a bio-inspired basal ganglia model within a foraging task. The present work does not implement a basal ganglia model, but rather a lightweight action-selection mechanism designed to reduce oscillatory behavior and improve behavioral stability during transfer to the three-dimensional simulator.

\subsection{Evolutionary procedure and final checkpoint}

The two agents co-evolve using a genetic algorithm with elite selection. The 10 best individuals from each generation pass directly to the next generation, while the parents for crossover are randomly selected from a pool of the 25 fittest individuals. The algorithm also employs uniform crossover and Gaussian mutation (probability 0.2, noise $N(0,0.12^2)$), with a population of 32 individuals per role and repeated evaluation (\emph{eval\_repeats}=3) to reduce the noise of the fitness signal under the stochastic curriculum of the environment. The official checkpoint used for the results of this work (\emph{v15\_full\_final}) was obtained by loading the weights from a previous run (\emph{social3d\_v10\_nn\_annealing\_gen143.json}) and continuing evolution for an additional 150 generations. During this subsequent 3D evolutionary phase, the following parameters were applied jointly: SOCIAL\_RESIDUAL\_BOOST=3.0, SOCIAL\_RESCUE\_BONUS=20,000, recruit\_prob=0.90, and a recalibration of the front-unlocking bonuses (\emph{FRONTAL\_UNBLOCK\_BONUS}=100, \emph{FRONTAL\_FORWARD\_PENALTY}=-50). These parameters were retained from prior development iterations of the system and were applied consistently throughout the additional evolutionary phase.

\texttt{FRONTAL\_UNBLOCK\_BONUS} and \texttt{FRONTAL\_FORWARD\_PENALTY} are fitness incentive terms inherited from the 2D implementation, active only when the agent detects a nearby obstacle in front (the same wall-proximity criterion used by other parts of the controller). The first term rewards the agent when it manages to move away from the obstacle, whereas the second penalizes it for continuing to move toward it. In the official checkpoint, both values were recalibrated from 50/-25 to 100/-50, matching the absolute magnitude used in the 2D engine; the 2:1 ratio between the bonus and the penalty already matched between both engines before this recalibration.

The activation criterion for the stall detection mechanism (\emph{is\_stuck}, see Section~\ref{sec:stall_criterion}) remained unchanged during the training of this checkpoint. The deactivation criterion, however, differed between the training and evaluation stages. The weights of \emph{v15\_full\_final} were trained using the original strict criterion, which required a perfect streak of 30 consecutive movement ticks to exit the stalled state. In contrast, all evaluations reported in this work were performed using the tolerant window criterion described above, which considers the agent freed when 24 of the last 30 ticks correspond to effective movement.

For the analysis of weight bias between agents (Section~\ref{sec:results}), a
distinction is made between inherited evolutionary runs starting from
the same baseline checkpoint and independent runs, initiated from
scratch with their own initialization. Only the latter are considered
statistically independent samples; inherited runs are reported as
complementary qualitative evidence, but are excluded from formal
hypothesis tests because they do not follow an initial independence
condition.

\subsection{Diagnostic and analysis tools}

The script \emph{compare\_weights\_ab.py} performs a direct comparison of the trained weights
and without executing any episodes, the magnitude of the weight
vectors measured by the L2 norm (Euclidean length) associated with each
action column in w\_out and w\_res between the two neural
networks \emph{brain\_a} and \emph{brain\_b}, focusing on the turning
columns (left\_turn, right\_turn).

On the other hand, the script \emph{diagnose\_signal\_effect\_windowed3d.py} (3D engine) and its adaptation to the 2D engine (\emph{diagnose\_signal\_effect\_windowed2d.py}) examine, within real-world episodes, the decisions made with a social cue heard versus those made without a cue. This is done using two metrics: net approach to the partner and stall-resolution rate.
By means of a flag (\emph{-\/-reverse\_order}), passed via command line,
the processing order between agents within the decision cycle was
reversed. This was used as a methodological control to evaluate whether the observed asymmetries between \emph{brain\_a} and \emph{brain\_b} could be explained by the physics engine's computational order. Furthermore, \emph{caracterizacion\_baseline\_v15.sh} executes the full battery of diagnostics reported in Section~\ref{sec:results} on a fixed checkpoint. This battery includes three measurements: the success rate with and without signal (n=300), the stalled time (n=150), and the stall-resolution effect (n=150), each evaluated on an independent sample of episodes. The
script allows parameterizing the step budget for each episode in each
run.

For the purposes of the stall-resolution analysis, n=150 indicates complete episodes rather than analyzed cases. Each episode generates many decision points, some with a signal present and others without. Therefore, the actual number of \emph{stall-with-signal} and \emph{stall-without-signal} cases compared in Table~\ref{tab:stall_resolution} and Figure~\ref{fig:unsticking_effect} (Section~\ref{sec:3.3}) is substantially larger than 150 and varies depending on the evaluated time budget.

\section{Results}
\label{sec:results}

\subsection{Performance baseline and social signal effect}
\label{sec:3.1}

For the final checkpoint (\emph{v15\_full\_final}, episode budget of 6,000 steps, $n=500$), the success rate of agent A was 49.8\% with the social signal enabled and 49.0\% with the signal disabled (a difference of -0.8 percentage points), whereas agent B achieved 71.2\% with the signal enabled and 73.4\% with the signal disabled (a difference of +2.2 percentage points). In neither agent was the difference between the signal-enabled and signal-disabled conditions distinguishable from sampling noise. This pattern is consistent with previous exploratory measurements obtained using a smaller sample ($n=300$), indicating that the observed result is stable with respect to sample size. The performance asymmetry between agents (B outperforming A in both conditions) remained stable throughout the experiments. Potential factors associated with this asymmetry are examined in Section~\ref{sec:3.4}.

\subsection{Effect of the time budget on performance}
\label{sec:3.2}

\begin{figure}[!ht] \centering \includegraphics[width=0.8\textwidth]{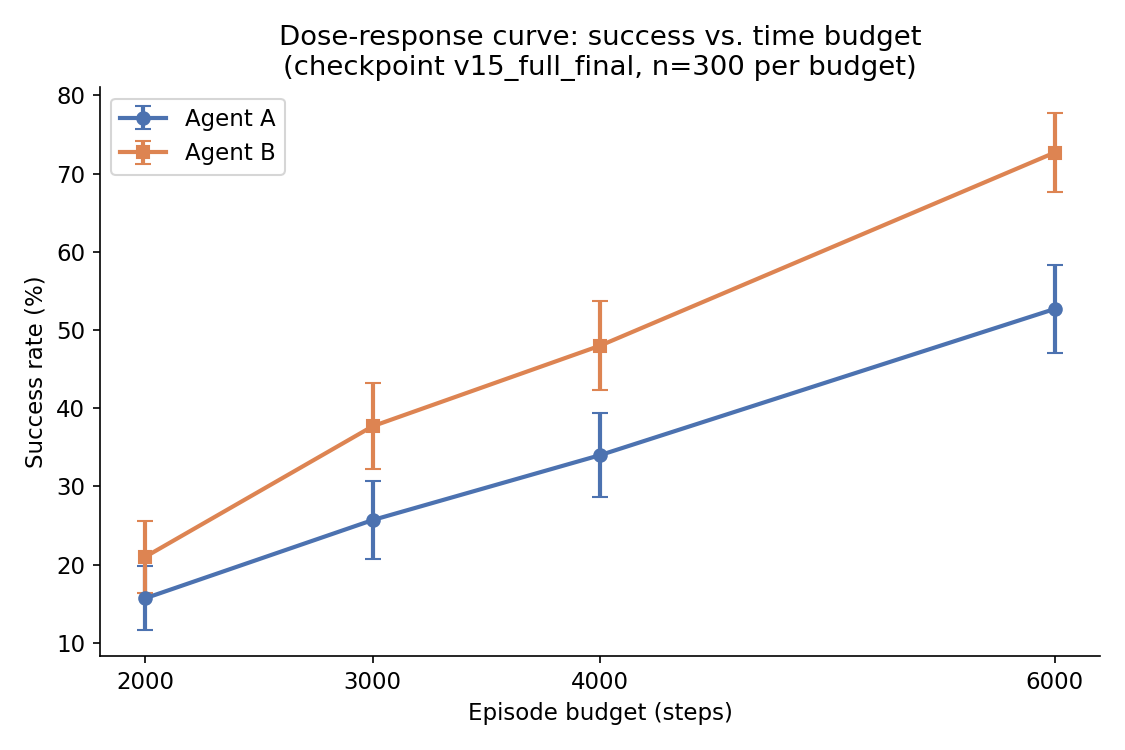} \caption{ Success rate of agents A and B as a function of the episode time budget (2,000 to 6,000 physics steps) for checkpoint \emph{v15\_full\_final} without retraining. Error bars represent 95\% confidence intervals for a binomial proportion ($n=300$ episodes per budget). The success rate increased for both agents as the time budget increased. Agent B maintained a higher success rate than agent A across all evaluated budgets. } \label{fig:dose_response} \end{figure}

The time budget allocated to each episode (\emph{max\_steps}) was strongly associated with the observed performance.
The same checkpoint (v15\_full\_final, without retraining) was evaluated under
four different budgets (2,000, 3,000, 4,000, and 6,000 physics steps),
with $n=300$ episodes for each budget. The success rate of both agents increased approximately linearly across the evaluated range, with no evidence of a plateau within the tested budgets: Agent A increased from 15.7\% to 52.7\%, and
Agent B from 21.0\% to 72.7\%, between the extremes of the range. The
observed growth rate, calculated between consecutive points on the
curve, remained approximately between 8 and 17 percentage points of
success for every additional 1,000 steps (Figure~\ref{fig:dose_response}).

Since performance did not stabilize within the explored range, 6,000 steps per episode were used for the remaining analyses. Within the evaluated range, this budget provided the highest observed success rates and was therefore adopted as the reference condition for subsequent experiments.

\subsection{Effect of social signaling on stall resolution}
\label{sec:3.3}

\begin{figure}[!ht] \centering \includegraphics[width=0.8\textwidth]{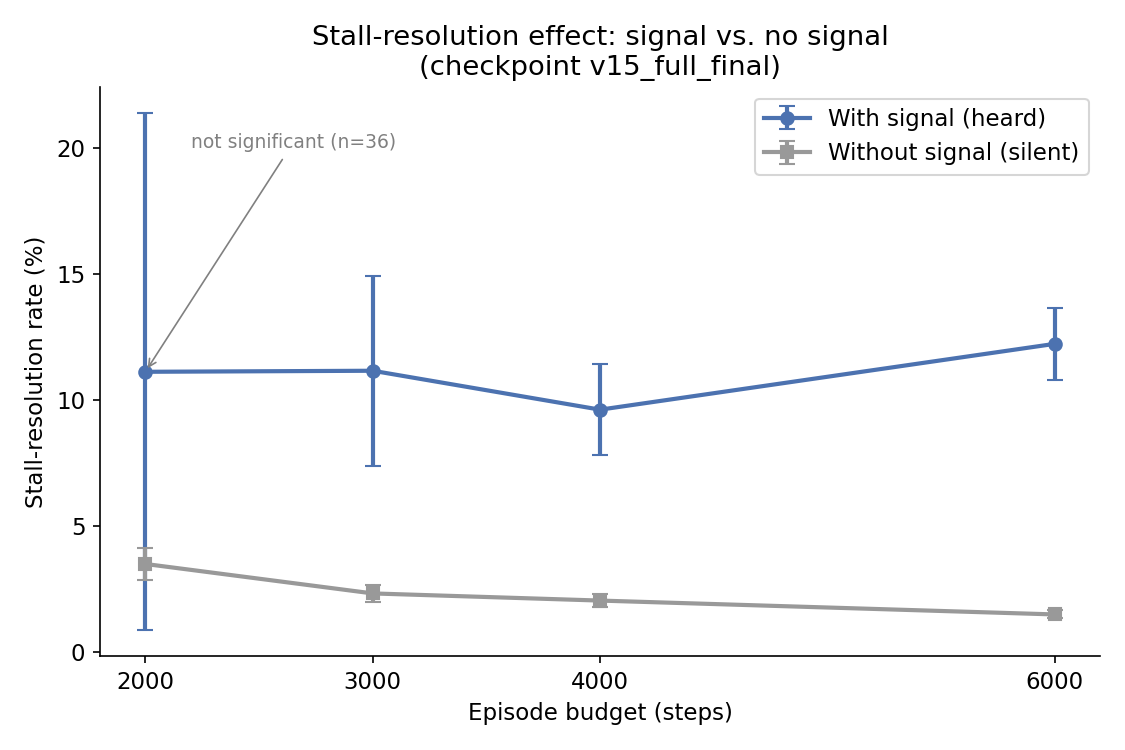} \caption{ Stall resolution rate with social signal present (heard) and absent (silent), as a function of the episode time budget. Error bars show the 95\% confidence interval for a binomial proportion, calculated from the counts observed in each condition. For the 2,000-step budget, the observed difference did not reach statistical significance, coinciding with the small number of decision points classified as stalled with a signal present ($n=36$). } \label{fig:unsticking_effect} \end{figure}

Within the evaluated episodes, the rate of resolution of stalled situations was compared between decision points in which the agent received a social signal from its partner and decision points in which no signal was present, for the
same four time budgets. The resolution rate was consistently higher in
the presence of a signal (Table~\ref{tab:stall_resolution}, Figure~\ref{fig:unsticking_effect}). The effect was
statistically significant in the budgets of 3,000 steps (difference of
8.8 percentage points, z=4.58, $p<0.0001$), 4,000 steps
(difference of 7.6 percentage points, z=8.13, $p<0.0001$), and
6,000 steps (difference of 10.7 percentage points, z=14.77,
$p<0.0001$). In the shortest budget (2,000 steps), the observed
difference (7.6 percentage points) did not reach statistical
significance (z=1.45, p=0.146). This result coincided with the small number of stalled decision points with a signal present in that condition ($n=36$), which limited the statistical power of the comparison.

\begin{table}[!ht] \centering \caption{ Stall resolution rate with and without social signal, by episode time budget. } \label{tab:stall_resolution} \begin{tabular}{cccccc} \hline Budget & With signal & Without signal & Difference & n & p \\ (steps) & (\%) & (\%) & (pp) & (with signal) & \\ \hline 2,000 & 11.1 & 3.5 & 7.6 & 36 & $0.146$ \\ 3,000 & 11.2 & 2.3 & 8.8 & 269 & $<0.0001$ \\ 4,000 & 9.6 & 2.0 & 7.6 & 1,020 & $<0.0001$ \\ 6,000 & 12.2 & 1.5 & 10.7 & 2,054 & $<0.0001$ \\ \hline \end{tabular} \end{table}

These results indicate that social signaling was consistently associated with higher stall-resolution rates. In conjunction with the results from Section~\ref{sec:3.1}, this pattern is more consistent with a local assistance function than with a general navigation mechanism for guiding an agent toward its partner.

\subsection{Weight asymmetry between agents}
\label{sec:3.4}

We examined whether the performance asymmetry between agents was associated with measurable differences in the trained weights, comparing the weights associated with the right\_turn action between brains A and B
(w\_out and w\_res), in eight evolutionary runs of mutually independent
origin (four in the 3D simulator, four in the 2D simulator). Lineages derived by
successive inheritance from the same checkpoint of origin were excluded
from the statistical analysis because they did not constitute
independent samples, although they are mentioned later as complementary
qualitative evidence.

\begin{figure}[!ht] \centering \includegraphics[width=0.8\textwidth]{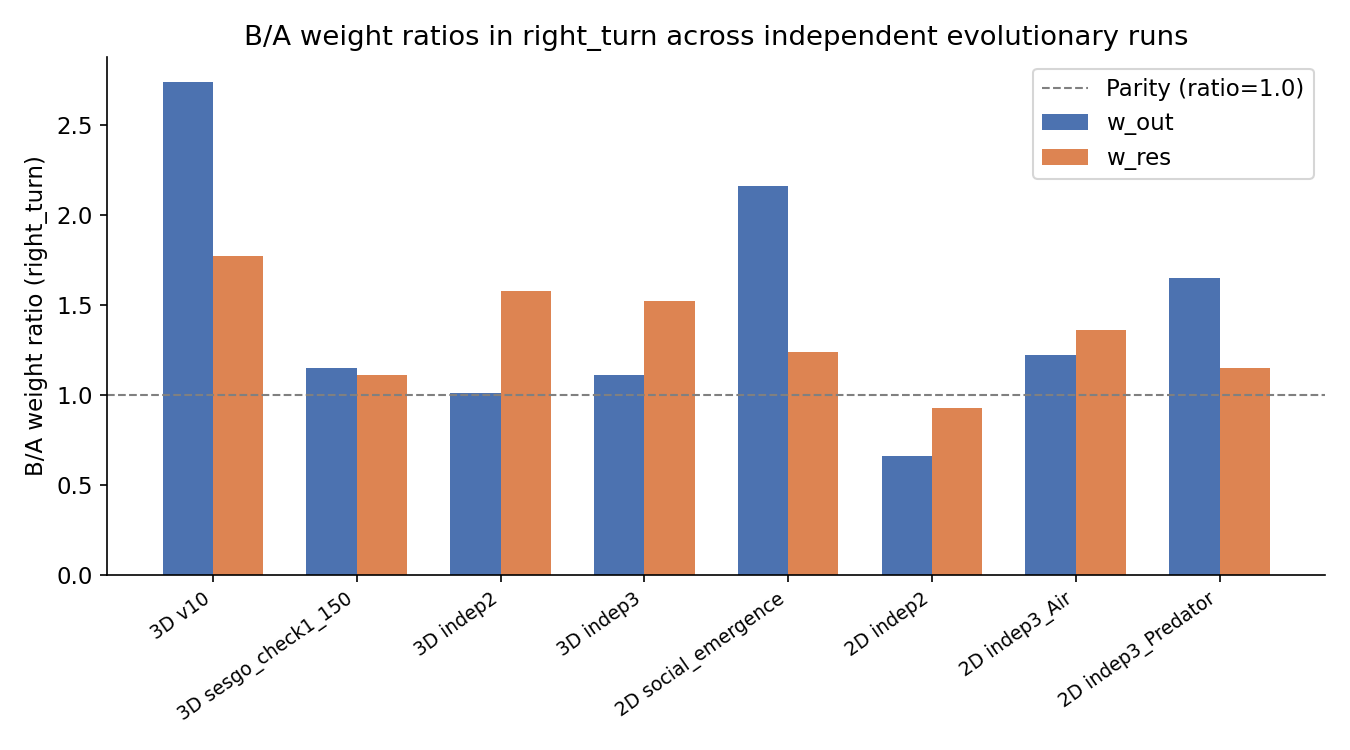} \caption{ B/A weight ratios in the \texttt{right\_turn} action column (\textit{w\_out} and \textit{w\_res}) for eight independent evolutionary runs (four in the 3D simulator and four in the 2D simulator). The horizontal dashed line indicates parity between agents (ratio = 1.0). The \textit{w\_res} ratio was significantly above parity ($p=0.012$), whereas the \textit{w\_out} ratio showed the same trend without reaching conventional statistical significance ($p=0.098$). The labels on the x-axis correspond to the original checkpoint names of the evolutionary runs and are retained for traceability. } \label{fig:weight_ratios} \end{figure}

The B/A weight ratio in w\_res was significantly greater than 1.0 (mean
= 1.33, SD = 0.28, t(7) = 3.38, p = 0.012, 95\% CI [1.10, 1.57]).
The same ratio in w\_out showed a trend in the same direction, without
reaching conventional significance (mean = 1.46, SD = 0.69, t(7) = 1.91,
p = 0.098, 95\% CI [0.89, 2.03]) (Figure~\ref{fig:weight_ratios}). Overall, the results showed a consistent trend toward values above parity, although with considerable variability among independent runs.
Most independent runs favored agent B in the same direction observed in the inherited lineages.

As a methodological control, it was examined whether this pattern could
be attributed to an artifact of the physics engine. The reversal of the
agent processing order within the decision cycle
(\emph{-\/-reverse\_order}) did not produce differences in the relative
performance of A and B, both in the 3D simulator (n=300 episodes) and in
the 2D simulator (n=300 episodes), suggesting that the order of computation, body creation, and initial state per agent are unlikely to account for the observed asymmetry.

\section{Discussion}
\label{sec:discussion}

\subsection{Summary of findings}

The results presented in Section~\ref{sec:results} show three related phenomena. First, the episode time budget was strongly associated with the observed performance. The success rate of both agents increased approximately linearly across the explored range, with no evidence of a plateau within the evaluated budgets, suggesting that evaluations using shorter episodes may underestimate the performance of the trained controllers. Second, in both the 2D and 3D simulators, the social communication mechanism was more consistent with a stall-assistance function than with a general navigation mechanism for guiding an agent toward its partner. This behavior was observed in both simulators. Third, a consistent asymmetry was observed between the trained weights of the two co-evolved roles. The results suggest that this asymmetry may contribute to the persistent performance differences observed between agents, although its precise origin remains unresolved.

\subsection{Same mechanism, different consequences depending on the body}

A common finding across all three results is that the same learned
mechanism may be irrelevant in one environment and costly in another.
For example, the weight bias toward right turns in agent B has no
measurable consequences in the 2D simulator. Both agents achieve a
near-ceiling success rate and show very little difference between them.
In contrast, in the 3D simulator, the same bias is associated with episodes of sustained one-way turns and prolonged stalling. This reduces the success rate of the agent exhibiting it. One possible interpretation is that the underlying mechanism does not change between simulators. In 2D, turning inefficiencies have little effect. In 3D, these same inefficiencies are more strongly associated with reduced performance. This contrast is consistent with the type of gap between simulation and physical transfer already documented in evolutionary
robotics \cite{jakobiNoiseRealityGap1995, trianniSelforganisationCommunicationGroups2006}, where it is observed that a controller can exploit regularities in its training environment that do not hold true
under more demanding physical conditions, without this implying a
failure of the controller itself.

The same idea applies to the communication mechanism. The stall-resolution rate per social signal was higher in 2D than in 3D. This difference does not necessarily imply that the communication mechanism itself changed between environments. Rather, the same mechanism operates under different physical constraints, which may alter its practical effectiveness. A controller that performs near perfectly in the 2D environment may therefore exhibit lower performance after transfer to the 3D simulator, even if the underlying communication strategy remains largely unchanged.

The above reformulates the central question of this work: why does a
performance ceiling persist in 3D? One possibility is a limitation in
the controller's architecture. Another is a limitation
in evolutionary incentives. In the latter case, fitness pressure would
have been insufficient for behaviors to emerge that were never necessary
in the 2D environment. The results of this work do not allow us to
definitively answer this question, but they do help to narrow it down.
The results of this work do not allow a definitive distinction between architectural and evolutionary explanations. However, the observed weight asymmetry suggests that evolutionary dynamics may play an important role in the emergence of persistent differences between the two co-evolved agents. An additional possibility is related to the co-evolutionary setup itself. Although both agents share the same body morphology, \emph{brain\_a} and \emph{brain\_b}  are evolved independently and maintain fixed roles throughout evolution. This combination of homogeneous bodies, heterogeneous controllers, and fixed roles may contribute to transfer vulnerability, although this interpretation remains speculative and was not directly tested in the present work.
One possible explanation is that small differences arising early during evolution were progressively amplified through selection. Furthermore, the observed asymmetry was not consistently reproduced by simply exchanging roles or processing order, suggesting that the phenomenon cannot be attributed solely to a fixed architectural property.

\subsection{Interpretation of the observed weight bias}

The observed bias pattern does not appear to be strictly deterministic, as it does not appear in all
independent runs with the same magnitude, nor is it explainable solely
by chance. Most independent runs reproduce the pattern in the same
direction as the inherited lineages. Furthermore, the magnitude of the
bias tends to increase in lineages that have accumulated more
generations of selection throughout their evolutionary history.

This pattern is reminiscent of producer biases described in the
literature on the evolution of communication, where an initially small
asymmetry between roles can be amplified by evolutionary dynamics until
it consolidates into a stable specialization of roles \cite{mirolliHowProducerBiases2008}. In a control run started from scratch (\emph{bias\_check1}), the pattern was not observed at 60 generations. However, when extended to 150 generations, the pattern appeared weakly. This result is consistent with the possibility that an early symmetry-breaking event becomes progressively reinforced through cumulative selection.

This question remains unanswered by this work. Distinguishing between these hypotheses would require a direct analysis of the controller's internal dynamics. For example, recurrent-state analyses similar to those proposed by Sussillo and Barak \cite{sussilloOpeningBlackBox2013} could be used to determine whether the observed asymmetry is associated with distinct dynamical regimes within the controller. This possibility is left for future work and is outside the scope of the present study.

The latter interpretation is also consistent with previous work in
evolutionary robotics. Previous studies have shown that the level of
selection and the genetic relationship between agents can influence role
specialization and the patterns of cooperation that emerge during
evolution \cite{floreanoEvolutionaryConditionsEmergence2007, waibelGeneticTeamComposition2009}. 
In this work, the fitness coupling between \emph{brain\_a} and \emph{brain\_b} is indirect and weak, as it
is limited to the shared social bonus. Under these conditions, role
differentiation may emerge as a trend, although not necessarily
uniformly. This aligns with the observed results; the eight independent runs show a trend toward differentiation, although the magnitude of the effect varies across lineages.

\subsection{Limitations}

Several limitations restrict the scope of these conclusions. First,
although the weight bias analysis included four independent runs on the
2D engine, the other three diagnostics of that engine (the stall-resolution effect, 
the turning pattern, and the processing order control) were
performed earlier, on the only 2D checkpoint available at that time
(\emph{social2d\_emergence.json}). The additional independent runs were generated at a later stage, specifically for weight bias analysis, and were not used to repeat those three diagnostics. Repeating those diagnostics on the additional independent 2D runs would be desirable and may provide a more robust estimate of their variability, but was outside the scope of the present study. Consequently, the comparison between simulators in those three aspects is qualitative and relies on a single checkpoint per simulator, not on replicated validation in both environments.  Second, the improvement applied to the official
checkpoint (v15) measurably strengthened the stall-resolution mechanism.
However, the 3D stall frequency remains much higher than in 2D, where
this phenomenon is infrequent. As a result, the observed improvement in
the stall-resolution mechanism was not sufficient to close the performance gap
between the two simulators. Third, the performance curve did not show a
plateau within the evaluated range (2,000 to 6,000 steps). The search
was not extended beyond that range because the objective was to
establish a reasonable budget for the task, not to find the maximum
possible performance. For this reason, the performance observed at 6,000
steps does not necessarily represent the upper limit achievable by
current controllers. 
Fourth, the statistical analysis of weight bias is
based on eight independent runs. This sample was sufficient to detect a
trend, but it is limited for accurately estimating its magnitude or
characterizing its distribution.

\subsection{Future work}

Two lines of inquiry stem directly from these results. The first is the analysis of the controller's internal dynamics. Its objective is to determine whether the observed weight asymmetry reflects a structural property of the controller or an asymmetry acquired during evolution. The absence of clear role-specific dynamics would be as informative as their presence, as it would strengthen the hypothesis that the phenomenon is associated with evolutionary dynamics rather than with a fixed architectural limitation. The second is to review the activation criterion for \emph{is\_stuck}, which remained unchanged from the original formulation (25 consecutive steps without effective displacement). This criterion determines when an agent enters the stalled state and may contribute to the higher stall frequency observed in 3D relative to 2D. In contrast, the deactivation criterion was revised and updated to the tolerant window described in Section~\ref{sec:stall_criterion} before the evaluations reported in this work.

A possible extension of this line of work is the analysis of internal controller representations using approaches related to latent-state modeling and learned world models \cite{hafnerDreamControlLearning2019, haWorldModels2018}. Such methods could help characterize whether the observed asymmetries are associated with distinct internal representations or decision regimes. This direction remains outside the scope of the present study.

\section*{Acknowledgements} The author gratefully acknowledges the support of Universidad Veracruzana. 
\section*{Funding Information} This work was supported in part by the Mexican National System of Researchers (SNII). Fernando Montes-Gonzalez is a member of the SNII (Researcher ID 30026). 
\section*{Author Contribution} The author confirms the sole responsibility for the conception of the study, methodology development, experimental implementation, analysis of results, and manuscript preparation. 
\section*{Conflict of Interest} The author states no conflict of interest. 
\section*{Ethical Approval} The conducted research is not related to either human or animal use. 
\section*{Data Availability Statement} 
The source code used for the simulations, evolutionary training, and diagnostics reported in this work is publicly available in the GitHub repository: \href{https://github.com/ferdiex/essim2d3d}
{https://github.com/ferdiex/essim2d3d}. 
An archived and citable version with a permanent DOI is available through Zenodo: 
\href{https://doi.org/10.5281/zenodo.21967055}
{https://doi.org/10.5281/zenodo.21967055}.
The raw data supporting the figures and tables reported in this study are available from the author upon reasonable request.

\bibliographystyle{IEEEtran} \bibliography{2Dto3D}

\end{document}